%% file: preprint.tex
\documentclass[letterpaper,11pt]{article}

\usepackage[utf8]{inputenc}
\usepackage[T1]{fontenc}
\usepackage{times}
\usepackage[margin=1in,top=0.45in]{geometry}
\usepackage{microtype}
\PassOptionsToPackage{numbers, sort&compress}{natbib} 

\usepackage{amsmath,amssymb,amsfonts,amsthm}

\usepackage{mathrsfs}
\usepackage{dsfont}
\usepackage{soul}

\usepackage{graphicx}
\usepackage[export]{adjustbox}
\usepackage{tabularx}
\usepackage{booktabs}
\usepackage{multirow}
\usepackage{makecell}
\usepackage{subfigure}
\usepackage{longtable}
\usepackage{wrapfig}
\usepackage{algorithm}
\usepackage{algpseudocode}

\usepackage{enumitem}
\usepackage{xspace}
\usepackage{blindtext}
\usepackage[most]{tcolorbox}
\usepackage{xcolor}
\usepackage[table]{xcolor}
\usepackage{etoc}
\usepackage{todonotes}

\usepackage{cite}
\usepackage{natbib}
\usepackage{url}
\usepackage[colorlinks=true,linkcolor=cyan,citecolor=cyan,filecolor=magenta,urlcolor=blue]{hyperref}
\usepackage[nameinlink,capitalise]{cleveref}

\hypersetup{
  pdftitle={LatentUMM: Dual Latent Alignment for Unified Multimodal Models},
  pdfauthor={Yinyi Luo, Wenwen Wang, Hayes Bai, Marios Savvides, Jindong Wang}
}

\theoremstyle{definition}

\renewcommand{\thefootnote}{\textasteriskcentered}

\setlist{leftmargin=5mm}

\definecolor{abstractbg}{RGB}{230,242,250}
\definecolor{abstractborder}{RGB}{200,210,220}
\usepackage{titlesec}

\titleformat{\section}
  {\sffamily\bfseries\Large} 
  {\thesection}              
  {1em}                      
  {}                         

\renewenvironment{abstract}
{
\begin{center}
\begin{tcolorbox}[
    colback=abstractbg,
    colframe=abstractborder,
    boxrule=0.6pt,
    arc=6pt,
    width=\textwidth,
    left=8pt,
    right=8pt,
    top=8pt,
    bottom=8pt
]
}
{
\end{tcolorbox}
\end{center}
}

\newcommand{\papertitle}{%
\sffamily\bfseries\fontsize{16}{1}\selectfont
LatentUMM: Dual Latent Alignment for Unified \\Multimodal Models
}

\newcommand{\method}{LatentUMM\xspace}

\newcommand{\paperauthors}{%
\sffamily
Yinyi Luo$^{1\ast}$, Wenwen Wang$^1$, Hayes Bai$^2$, Marios Savvides$^1$ and Jindong Wang$^{2\dag}$%
}

\newcommand{\paperdate}{$^1$Carnegie Mellon University \quad $^2$William \&\ Mary}

\begin{document}

\renewcommand{\thefootnote}{\fnsymbol{footnote}}
\setcounter{footnote}{0}
\thispagestyle{empty}

\noindent
\includegraphics[height=.6cm]{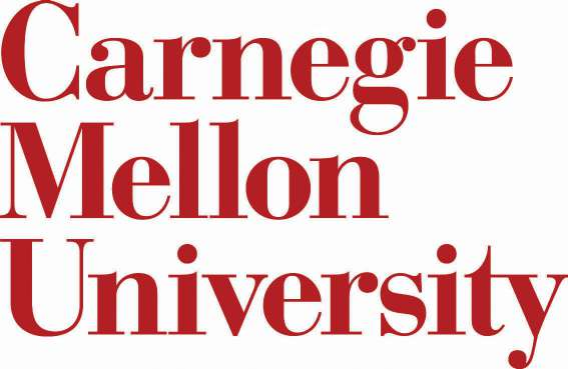} ~
\includegraphics[height=.6cm]{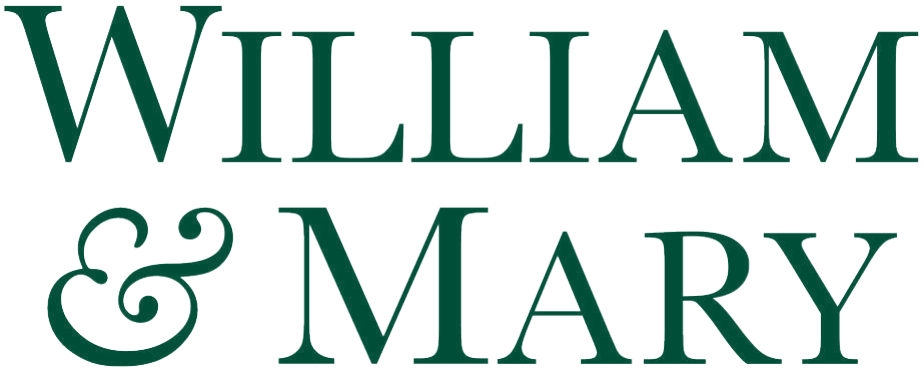}

\vspace{-.15in}

\noindent\rule{\textwidth}{0.8pt}

\vspace{0.65cm}

\begin{center}
    {\papertitle\par}
    \vspace{0.45cm}
    {\large \paperauthors\par}
    \vspace{0.2cm}
    {\normalsize \paperdate\par}
\end{center}

\footnotetext[1]{Contact: yinyil@andrew.cmu.edu}
\footnotetext[2]{Corresponding authors: jdw@wm.edu}

\begin{abstract}
  Unified multimodal models (UMMs) achieve strong performance in both understanding and generation by learning a shared latent space, yet they often exhibit functional inconsistency between these two capabilities. We observe that this issue does not stem from a lack of shared representations, but from the absence of explicit alignment between the transformations that map into and out of the latent space. As a result, generation and re-encoding can follow inconsistent trajectories, leading to semantic drift under modality transitions. In this work, we propose \textbf{\method}, a framework that constructs an enhanced shared latent space to explicitly align these transformations and improve cross-modal consistency. \method consists of two stages. First, dual latent alignment enforces consistency at both the modality and capacity levels: cross-modal alignment uses a stronger embedding model to impose structured cross-modal semantics, while dual capacity alignment enforces bidirectional consistency under generation and re-encoding. Second, latent dynamics stabilization improves robustness via stochastic latent rollouts and preference optimization, favoring trajectories that better preserve semantic consistency. 
  Experiments show that \method consistently improves multimodal consistency across diverse architectures.
 Code is available at: \url{https://github.com/AIFrontierLab/TorchUMM/tree/main/src/umm/post_training/LatentUMM}.
\end{abstract}

\vspace{-10pt}
\section{Introduction}
\vspace{-5pt}

Unified multimodal models (UMMs) have emerged as a promising paradigm for integrating understanding and generation within a single architecture \citep{zhao2025unified, chen2025janus, deng2025emerging, xie2025show}.  
By jointly training on multimodal data, UMMs learn a shared latent space that supports both multimodal understanding and generation, enabling them to interpret inputs and produce outputs across modalities. 
Despite this progress, recent studies reveal a persistent inconsistency between these two capabilities \citep{yan2025can, wang2026quantifying, wen2026unig2u, su2026generation, mao2025unirl, yang2026pseudo}.

In particular, although UMMs can generate high-quality images conditioned on text, they often fail to maintain semantic consistency when processing their own outputs. 
For example, a model may generate an image that correctly reflects a textual prompt, yet produce mismatched or incomplete semantics when asked to reinterpret that same image \citep{jiang2026can, wang2026quantifying, han2025turning}.  
This discrepancy suggests that joint training primarily aligns representations at the feature level, but does not guarantee consistency between the functional behaviors of understanding and generation.

We argue that this limitation arises because sharing a latent space is not sufficient to align how the model uses it.  
More fundamentally, learning a semantically consistent and cross-modal coherent latent space is highly non-trivial \citep{jiang2023understanding}.
Unlike explicit symbolic representations, latent spaces are learned implicitly and lack direct supervision on their structure, making them difficult to interpret, constrain, or verify \citep{yu2026latent,liu2019representation}.
As a result, semantic information is not necessarily preserved in a consistent manner across modalities, especially during transitions between understanding and generation.
Although both capabilities operate on the same latent representation, they rely on different mappings—understanding process encodes inputs into the latent space, while generation decodes latent representations into outputs \citep{zhu2024latentexplainer}.  
These mappings are learned implicitly during joint training and are not explicitly coordinated \citep{zhu2017unpaired, hafner2019learning, xiao2025omnibridge}.  
As a result, transitions across modalities can become inconsistent, leading to semantic drift when the model alternates between understanding and generation.
This phenomenon can be directly observed through a consistency diagnostic that measures semantic drift under repeated cross-modal transformations (details in Section \S\ref{sec-app-consistency}). 
Even when operating within a shared latent space, baseline UMMs exhibit progressively increasing deviation in latent representations across transformation steps, providing evidence of latent drift.

In response to this issue, a line of work on self-correction has been explored, including both inference-time refinement and post-training approaches, where models iteratively improve consistency by re-evaluating and revising their outputs \citep{han2025turning, luo2026self, jin2025srum, mao2025unirl, yang2025hermesflow}.  
While effective in improving empirical performance, these methods operate by iteratively correcting outputs or adapting model behavior, without explicitly constraining the underlying interaction between understanding and generation within the shared latent space.  
As a result, they do not address a key source of inconsistency, i.e., the lack of alignment between the bidirectional encoding and decoding processes.

\begin{figure*}[t!]
  \centering
  \vspace{-.1in}
  \includegraphics[width=.8\linewidth]{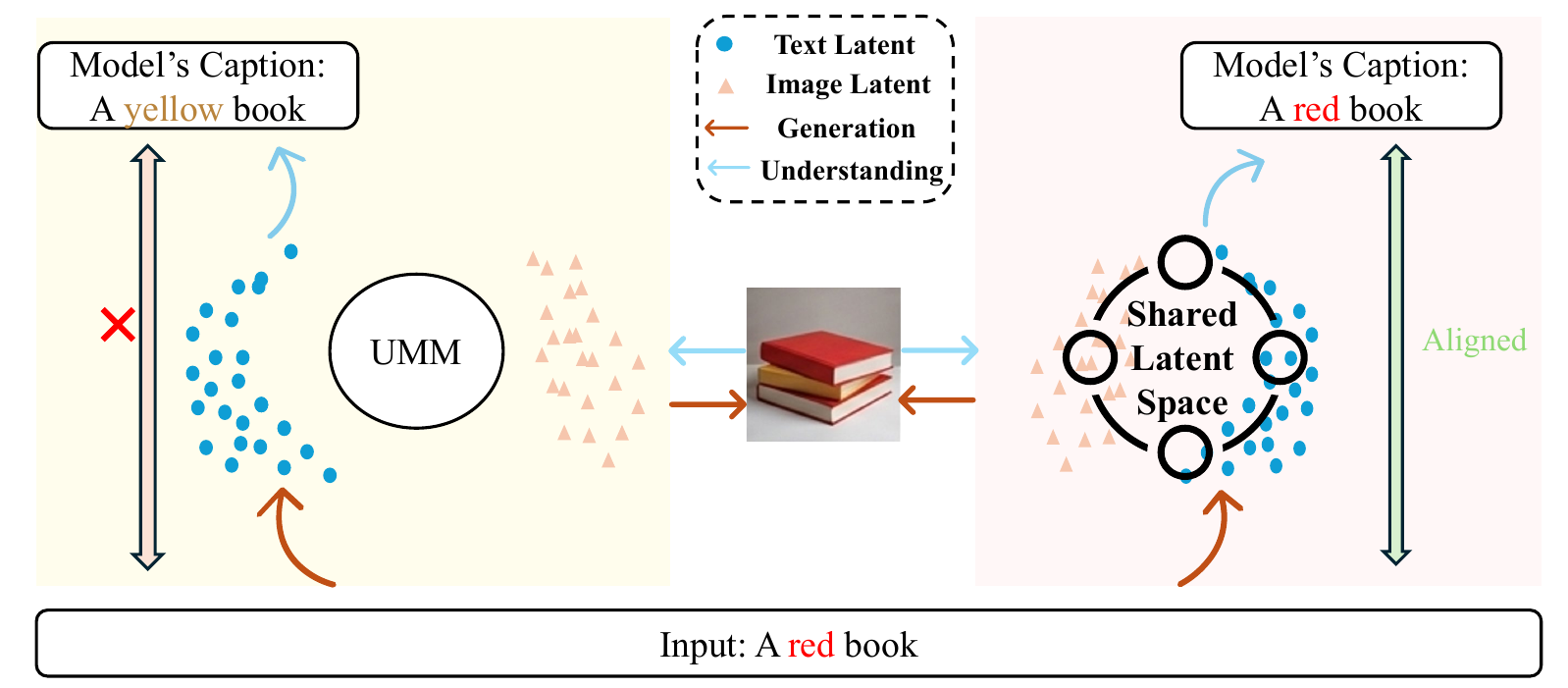}
  \vspace{-.1in}
  \caption{Overview of baseline UMMs (left) versus our proposed \method (right).}
  \label{fig:teaser}
  \vspace{-.2in}
\end{figure*}

To address this issue, we propose \textbf{\method}, a framework that enforces \textit{dual latent alignment} as a two-step process.
Instead of relying solely on a shared representation space learned via joint training, our approach explicitly couples the transformations of understanding and generation through the following key steps:
1) Dual Capacity Alignment.
We model both capabilities as bidirectional mappings within a shared latent space and enforce a structured alignment between their induced transformations.
This ensures that information remains consistent when transitioning across modalities, such that generated outputs can be reliably reinterpreted without semantic degradation.
As illustrated in Figure~\ref{fig:teaser}, while baseline UMMs exhibit divergence under modality loopback, \method maintains consistent structure and semantics through this alignment.
2) Latent Dynamics Stabilization.
However, enforcing alignment along a single transformation path may be insufficient in complex scenarios.
To further improve robustness and handle ambiguity, we introduce a rollout-based optimization mechanism in the latent space.
By perturbing latent representations, we sample multiple candidate transformation trajectories and evaluate their consistency, selecting more stable paths via preference optimization.
This allows the model to better handle ambiguity and avoid degenerate or inconsistent mappings before incorporating them into a post-training stage as an additional supervisory signal.

\textbf{Contributions.}
Our contributions are three-fold:
\begin{enumerate}[leftmargin=2em]
\setlength\itemsep{0em}

\vspace{-5pt}
\item \textbf{Latent alignment.}  
We propose \method that explicitly enforces alignment of both modalities and capacities in a shared latent space, beyond the original UMM latent space.

\item \textbf{Rollout-based latent optimization.}  
We introduce a rollout and preference optimization strategy that explores multiple latent transformation trajectories and selects semantically consistent ones, improving robustness and stability.

\item \textbf{Extensive experiments.}  
We demonstrate that \method is model-agnostic and consistently improves UMM consistency across diverse architectures.

\end{enumerate}

\section{Related Work}
\label{sec-related}
\vspace{-8pt}

\textbf{Unified Multimodal Models.}
Recent advances in UMMs aim to integrate multimodal understanding and generation within a single architecture \citep{zhao2025unified, yin2024survey,deng2025emerging, chen2025blip3, geyer2023tokenflow, wang2026deepgen}. A dominant approach adopts decoder-only autoregressive transformers trained on interleaved multimodal tokens \citep{cui2025emu3, wang2024emu3}. 
Another line of work explores hybrid generative frameworks that combine autoregressive modeling with diffusion or flow-based components \citep{xie2024show, wu2024janus, ma2024janusflow, chen2025janus}, which improve visual generation and cross-modal alignment. Additional efforts investigate modular or lightweight designs that bridge multimodal LLMs and generative models through intermediate connectors \citep{wu2025openuni}. Despite rapid progress, existing approaches primarily focus on architectural unification and scaling, leaving deeper issues of cross-modal consistency and reasoning less explored.




\textbf{Inconsistency in UMMs.}
Recent work has begun to examine whether UMMs exhibit consistent behavior across understanding and generation \citep{huang2025interleaving, qin2025uni, jin2026latentum}.  
Despite sharing a common architecture and latent space, a growing body of evidence reveals a persistent inconsistency between these two capabilities.
In particular, models that generate high-quality outputs often fail to maintain semantic consistency when processing their own generations \citep{wang2026quantifying, wen2026unig2u, jiang2026can}.  
This indicates weak bidirectional coherence and suggests that current UMMs do not fully unify understanding and generation.
Prior work attributes this issue to the nature of existing training objectives.  
While modalities are aligned in a shared latent space through distribution-level supervision \citep{radford2021learning, li2023blip}, the mappings into and out of this space are not explicitly coordinated.  
As a result, understanding and generation remain loosely coupled and can exhibit inconsistent behavior \citep{wu2024janus, xie2024show}.
Recent benchmarks further highlight this limitation by evaluating consistency under modality loopback or cross-modal transformations \citep{wang2026quantifying, yan2025can, zou2025uni}.  
These results show that strong task performance does not guarantee consistent cross-modal behavior, motivating the need for explicit bidirectional alignment.

\vspace{-14pt}
\section{Method}
\vspace{-5pt}

\begin{figure*}[t!]
  \centering
  \includegraphics[width=\linewidth]{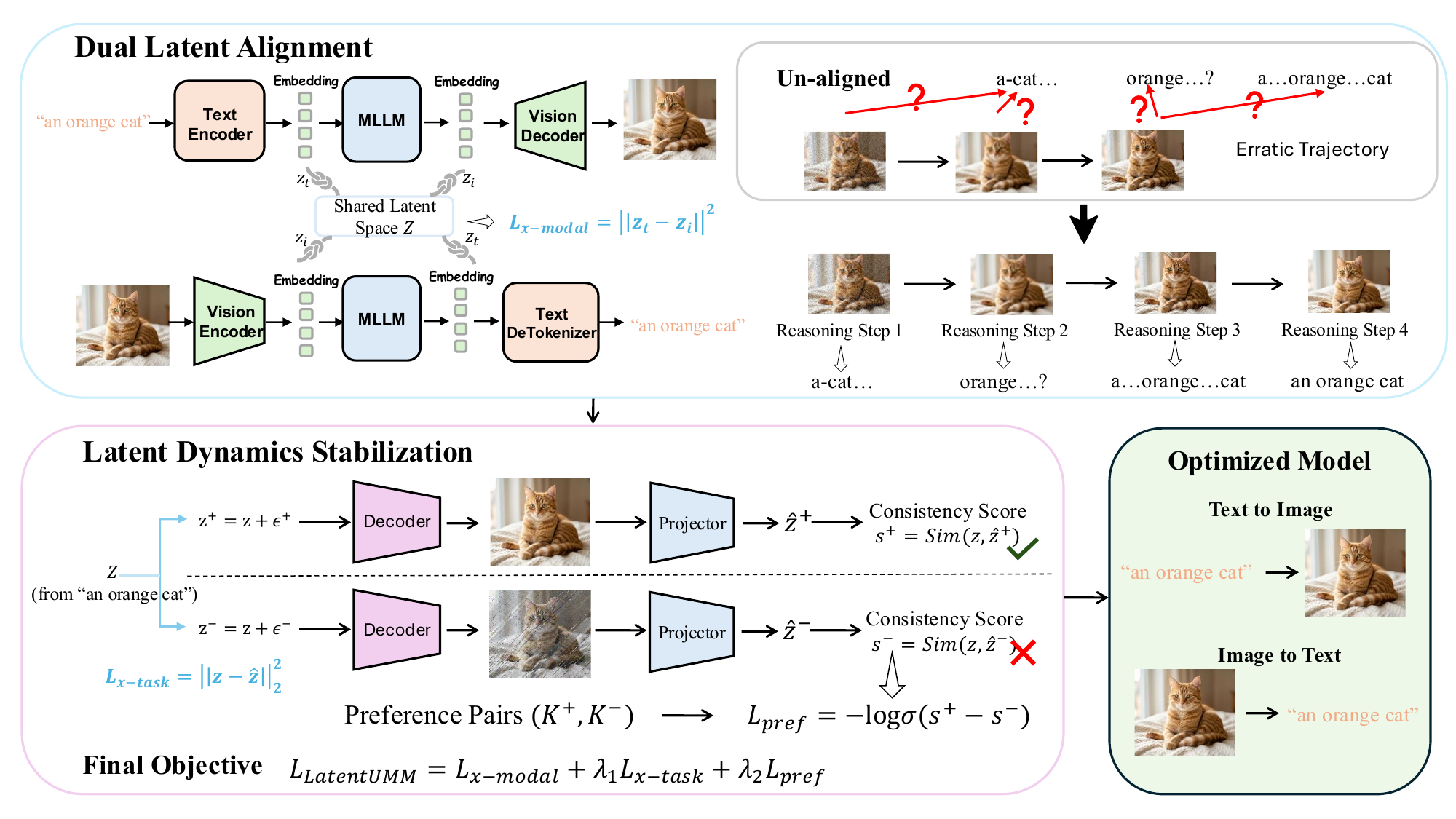}
  \vspace{-.1in}
  \caption{Overview of \method.}
  \label{fig:pipeline}
  \vspace{-.2in}
\end{figure*}

UMMs are typically trained to jointly model understanding and generation within a shared latent space. While this joint training encourages a degree of cross-modal alignment, we observe a persistent \textit{functional inconsistency} between the two capabilities: representations that support strong multimodal understanding do not always induce faithful generation, and conversely, generative trajectories often deviate from semantically consistent latent semantics \citep{rohrbach2018object, chen2021improving, ho2022classifier}.

This suggests that a single shared latent space optimized only under joint training is insufficient to guarantee cross-capability consistency. 
To address this limitation, we propose to construct an \textbf{enhanced shared latent space} that is explicitly refined using a stronger embedding model. The key idea is to further align the latent representations induced by understanding and generation, such that both functions become mutually consistent under the same semantic geometry.

\textbf{Overview.}
We propose \method, a framework that refines the latent space of a pretrained UMM into a more consistent multimodal representation space. As shown in \Cref{fig:pipeline}, \method operates as a two-step process: (1) Dual Latent Alignment leverages a stronger embedding model to regularize latent semantics and enforces consistency between understanding-induced and generation-induced transformations. (2) Latent Dynamics Stabilization then applies rollout-based preference optimization to handle complex scenarios, ensuring robust and stable latent trajectories.

\subsection{Problem Formulation}

Given paired text-image data $(x_t, x_i)$, where $x_t$ and $x_i$ denote text and image inputs respectively, we aim to learn a unified latent representation $z \in \mathcal{Z}$ in a shared latent space $\mathcal{Z}$.
We define modality-specific encoders $E_t(\cdot)$ and $E_i(\cdot)$ that map inputs into modality-specific latent embeddings: 
$z_t = E_t(x_t) \text{ and } z_i = E_i(x_i)$,
where we assume $z_t, z_i \in \mathbb{R}^d$, i.e., they lie in the same latent dimensionality $d$ for compatibility in subsequent fusion \citep{radford2021learning,faghri2017vse++,frome2013devise}.
We construct the unified latent representation via a fusion operator $F(\cdot,\cdot)$:
$z = F(z_t, z_i)$,
where $F$ is implemented by UMM’s latent fusion module (e.g., cross-attention or pooling), and is kept fixed during our refinement stage unless otherwise specified.
A decoder $G(\cdot)$ maps latent representations back to the image space:
$\hat{x} = G(z)$.

To construct a semantically more stable supervision signal over the same latent dimensionality, we use a stronger embedding model $E^{*}(\cdot)$ to map any input into a refined embedding space:
$\phi(x) = E^{*}(x) \in \mathbb{R}^d$,
where we explicitly assume that $E^{*}$ produces embeddings in the same dimensionality $d$, enabling direct geometric comparison (i.e, gemini embedding model \citep{lee2025gemini} projects all modalities into same dimension). 
Although dimensionality reduction or projection into a different space is common, it would require an additional learned mapping \citep{radford2021learning,chen2020simple}, potentially introducing optimization complexity and obscuring the source of alignment improvements. 
Our design isolates the role of the supervision signal itself.
For convenience, we define:
$\phi(\hat{x}) := E^{*}(\hat{x})$.
Here, $E^{*}$ is a fixed high-capacity model used only for supervision and alignment but not for inference.
\method aims to perform dual latent alignment to improve consistency, which will be introduced in next sections.

\vspace{-5pt}
\subsection{Dual Latent Alignment}

Dual latent alignment aims to achieve two levels of alignment: dual modal alignment between the visual and textual modalities, and dual capacity alignment for both the understanding and generation capacities of UMMs by modeling them in a shared latent space.
A stronger embedding model $E^*$ is used to induce both image and text representations into the same latent space that facilitates cross-modal alignment explicitly.
Then, we can perform capability alignment to improve the consistency.
 
\textbf{Dual modal alignment.}
We first align paired modalities in the embedding space induced by $E^{*}$:
\begin{equation}
\mathcal{L}_{\text{x-modal}} = \| \phi(x_t) - \phi(x_i) \|_2^2.
\end{equation}
Importantly, $\phi(\cdot)$ produces representations in the same dimensional space as $z$, allowing direct geometric alignment between the UMM latent space and the refined embedding space.
This objective enforces cross-modal semantic alignment in a more structured embedding space than the original UMM latent space, effectively inducing a refined shared semantic geometry.
In real experiments, the embedding model $E^*$ has many options such as Gemini Embedding model~\citep{lee2025gemini} and SigLIP~\citep{zhai2023sigmoid}, whose effectiveness is ablated in \S\ref{sec-exp-abla-latent}.

\textbf{Dual capacity alignment.}
Based on the alignment in shared space, to further improve consistency between understanding and generation, we define a bidirectional process in latent space.
Starting from a unified latent representation $z$, we generate an output $\hat{z} = \phi(\hat{x})$ and re-encode it into the refined embedding space:
$z \xrightarrow{G} \hat{x}$.
We then enforce latent bidirectional alignment to ensure that generation and re-encoding preserve semantic identity in the refined latent space:
\begin{equation}
\label{eq-cycle}
\mathcal{L}_{\text{x-task}} = \| z - \hat{z} \|_2^2.
\end{equation}

\vspace{-8pt}
\subsection{Latent Dynamics Stabilization}

However, these latent alignments are only performed at instance level, lacking distributional robustness for diverse samples in real-world applications.
Latent dynamics stabilization is further performed to improve the robustness.
The core idea is to leverage stochastic latent rollouts to generate perturbations, which will then be used to compute the similarity of each trajectory.
However, stochastic rollouts inevitably introduce high-variance perturbations, resulting in latent trajectories with varying semantic fidelity. 
While some trajectories preserve the underlying semantics, others may drift due to compounding generation and re-encoding errors. 
This naturally induces a relative ranking over trajectories based on their semantic consistency.
A straightforward approach would be to directly regress on these scores or average across trajectories. 
However, such objectives are sensitive to noisy or low-quality samples and may blur the semantic structure of the latent space. 
Instead, we formulate latent dynamics stabilization as a preference learning problem, which focuses on distinguishing more consistent trajectories from less consistent ones. 
This relative supervision provides a more robust and discriminative training signal under stochastic perturbations.

\textbf{Stochastic latent rollouts.}
We first sample stochastic perturbations around each latent representation:
\begin{equation}
z^{(k)} = z + \epsilon^{(k)}, \quad \epsilon^{(k)} \sim \mathcal{N}(0, \sigma^2 I), \quad k = 1, \dots, K,
\end{equation}
where $\epsilon^{(k)}$ denotes Gaussian noise sampled for the $k$-th rollout, $\sigma$ controls the perturbation magnitude (ablation in \S\ref{sec-app-hyperparamter}), $K$ is the number of sampled trajectories whose efficacy is studied in \S\ref{sec-exp-abla-rollout}.
Each trajectory follows:
$z^{(k)} \rightarrow \hat{x}^{(k)} = G(z^{(k)}) \rightarrow \hat{z}^{(k)} = \phi(\hat{x}^{(k)})$.
We then define a similarity function in the refined embedding space:
$\text{Sim}(a,b) = \frac{a^\top b}{\|a\|\|b\|}$.
As such, the self-consistency score of each trajectory is:
$s^{(k)} = \text{Sim}(z, \hat{z}^{(k)})$.

\textbf{Preference optimization.}
Following preference-based optimization~\citep{rafailov2023direct}, denote $\sigma(\cdot)$ as the sigmoid function, the preference loss is defined as:
\begin{equation}
\mathcal{L}_{\text{pref}} = - \log \sigma\big(s^{(k^+)} - s^{(k^-)}\big),
\end{equation}
where the preference pair $(k^+, k^-)$ is defined as:
\[
k^+ = \arg\max_k s^{(k)}, \quad k^- = \arg\min_k s^{(k)}.
\]

The final training objective for \method is formulated as:
\begin{equation}
\mathcal{L}_{\text{\method}} =
\mathcal{L}_{\text{x-modal}}
+ \lambda_{1} \mathcal{L}_{\text{x-task}}
+ \lambda_{2} \mathcal{L}_{\text{pref}},
\end{equation}
where $\lambda_1, \lambda_2 > 0$ are two trade-off hyperparameters that control the relative importance of dual consistency and preference optimization, respectively.

\method thus constructs a refined shared latent space that enforces consistency between understanding and generation via both cross-modal alignment and bidirectional-consistent latent dynamics. Unlike standard UMM training that relies solely on joint optimization, our method introduces an external embedding-driven geometric constraint that stabilizes multimodal reasoning trajectories.

\vspace{-10pt}
\section{Experiments}

\vspace{-5pt}
\subsection{Experimental Setup}

\textbf{Datasets.}
We perform training on Text-to-Image-2M dataset \citep{xie2025reconstruction}, a large-scale collection of paired text-image data.
We conduct evaluation across generation, understanding, and editing tasks.
Generation quality is measured using DPG-Bench \citep{hu2024ella}, U-Eval \citep{li2026ueval}, and WISE \citep{niu2025wise}. Understanding is evaluated on MME \citep{fu2023mme}, MMMU \citep{yue2024mmmu}, MMBench \citep{liu2024mmbench}, MathVista \citep{lu2023mathvista} and MM-Vet \citep{yu2023mm}.
Editing is evaluated on ImgEdit benchmark \citep{ye2025imgedit}.
Furthermore, we evaluate the consistency on Unified-Bench \citep{yan2025can} and RealUnify \citep{shi2025realunify}. 
All experiments are conducted using TorchUMM \citep{luo2026torchumm} for fair comparison.

\textbf{Backbones and Embedding Models.}
Bagel \citep{deng2025emerging} is selected as our main base model; Janus-Pro \citep{chen2025janus} and Harmon \citep{wu2025harmonizing} are adopted for flexibility.
All post-training methods, including SFT, RecA \citep{xie2025reconstruction}, UniGame \citep{su2025unigame} and UniCot \citep{qin2025uni}, are trained on the same Text-to-Image-2M dataset for fair comparison.
To enable alignment between text and image representations, we employ external embedding models to map both modalities into a shared latent space.  
Specifically, we extract embeddings for text and images using a pretrained encoder and project them into a unified representation space with the same dimensionality.  
By default, we use the Gemini embedding model \citep{lee2025gemini}.  
We further perform ablations with CLIP \citep{radford2021learning} and SigLIP \citep{zhai2023sigmoid}.

\vspace{-5pt}
\subsection{Main Results}

\input{tab/bagel_understanding}
\textbf{Improved understanding performance.}  
As shown in \Cref{tab:bagel-understanding},  
\method consistently improves understanding and outperforms post-training baselines on all benchmarks.
The improvements are most evident on comprehensive evaluation suites (MME and MMVet), where \method achieves the strongest gains, suggesting that latent consistency effectively enhances global multimodal alignment rather than overfitting to specific task formats. 
Notably, \method achieves the best performance on open-ended reasoning (MathVista Free-Form), indicating stronger capability in handling less structured, generative reasoning tasks. 
Compared to UniCot, whose chain-of-thought-style supervision appears to benefit structured reasoning benchmarks, \method’s latent consistency constraint better supports flexible reasoning that requires integrating generation and understanding.  

\input{tab/bagel_gen}

\textbf{Improved generation performance.}  
In \Cref{tab:bagel-generation}, \method consistently outperforms the baseline model on all generation benchmarks. 
On DPG-Bench, the gains are not only reflected in the overall score but are particularly pronounced in fine-grained dimensions such as entity and attribute generation, indicating improved precision in capturing object details and their properties. 
Notably, the largest relative improvement appears in the “Other” category, suggesting enhanced robustness in handling diverse or less structured generation scenarios that fall outside standard entity–relation patterns. 
On UEval, \method achieves a more significant gain in the image modality, indicating that latent consistency is especially beneficial for stabilizing visual generation, where errors can easily accumulate during the generative process. 
These results suggest that enforcing latent consistency improves both the fidelity of fine-grained content generation and the robustness of multimodal outputs, leading to more reliable and coherent generation across diverse scenarios.

\input{tab/editing}
\textbf{Improved editing performance.}  
\method also brings consistent improvements on the editing benchmark (\Cref{tab:bagel-editing}).  
On the overall setting, the gains are most pronounced in geometric mean, reflecting a balanced improvement across both semantic fidelity and visual quality. 
In particular, the increase in Semantic Correctness indicates that \method produces edits that better adhere to the intended semantic changes, while the improvement in Perceptual Quality shows that visual realism and perceptual consistency are preserved during editing. 
Overall, these results indicate that enforcing latent consistency enhances both the semantic accuracy and perceptual quality of edits, while improving their joint alignment. 
This leads to more stable and reliable editing outcomes, especially in scenarios requiring precise, localized modifications.

\vspace{-5pt}
\subsection{Ablation Study}
\label{sec:ablation}
\vspace{-5pt}

\subsubsection{Effect of Shared Latent Space and Embedding Models}
\label{sec-exp-abla-latent}
\vspace{-5pt}

\textbf{Shared Latent Space.}
We first evaluate the importance of shared latent space by comparing against a direct SFT baseline, which does not impose any alignment constraint. 
As shown in \Cref{tab:bagel-understanding,tab:bagel-generation}, while SFT remains competitive on certain metrics, it consistently underperforms alignment-based variants in overall performance.
This comparison highlights the benefit of explicitly modeling cross-modal consistency. By aligning generation and understanding within a shared representation space, the model achieves more coherent and transferable representations, leading to improved generalization across tasks. 
Moreover, \Cref{tab:ablation}(a) compares two strategies for leveraging the shared latent space.
Alignment within the existing shared latent space directly regularizes the pretrained latent representation, while constructing and enhanced shared latent space reshapes the latent geometry under the guidance of a stronger embedding model.
We observe that our method consistently achieves better performance, suggesting that the original latent space of the pretrained UMM is not sufficiently structured for reliable cross-modal alignment.
Simply enforcing alignment within this space is therefore limited by its inherent geometric imperfections.

\input{tab/ablation}

\textbf{Embedding Models.}
We then study the influence of different embedding models on latent alignment. 
\Cref{tab:ablation}(b) shows that replacing Gemini embedding with CLIP or SigLIP leads to slightly different performance trade-offs, and no single approach dominates. 
Nevertheless, Gemini embedding achieves the strongest overall performance, particularly on MMMU and MathVista.
These results suggest that embedding quality plays an important role in cross-modal alignment, especially for reasoning-intensive tasks. At the same time, the performance gap across embedding choices remains relatively small, indicating that \method is \textit{not} tightly coupled to a specific embedding model. This suggests that the gains primarily arise from the alignment mechanism rather than the embedding model.

\subsubsection{Effect of Rollout and Inference Design}
\label{sec-exp-abla-rollout}

We further analyze the impact of rollout depth, stochasticity, and decoding strategies. 

\textbf{Rollout steps.} Varying the rollout frequency leads to non-uniform effects across benchmarks. Shorter rollouts (e.g., $K=5$) improve certain metrics such as MME and MMMU, while longer rollouts (e.g., $K=20$) slightly benefit free-form reasoning on MathVista. However, no single configuration dominates, indicating a trade-off between stability and long-horizon aggregation (\Cref{tab:ablation}(c)).

\textbf{Noise scale.} Introducing moderate stochasticity improves performance compared to deterministic inference (\Cref{tab:ablation}(d)). In particular, a small noise level ($0.1$) provides a balance between exploration and stability, while excessive noise ($0.2$) begins to degrade on structured benchmarks.

\textbf{Decoding strategy.} More advanced decoding strategies further improve robustness (\Cref{tab:ablation}(e)). Both simple ensembling and self-consistency outperform single-pass decoding, with self-consistency yielding the most stable gains. However, the improvements remain incremental, suggesting that decoding primarily refines predictions rather than fundamentally altering model behavior.


\subsection{Analysis on Consistency Improvement}

\input{tab/unified}

\begin{figure*}[t!]
  \centering
  \includegraphics[width=\linewidth]{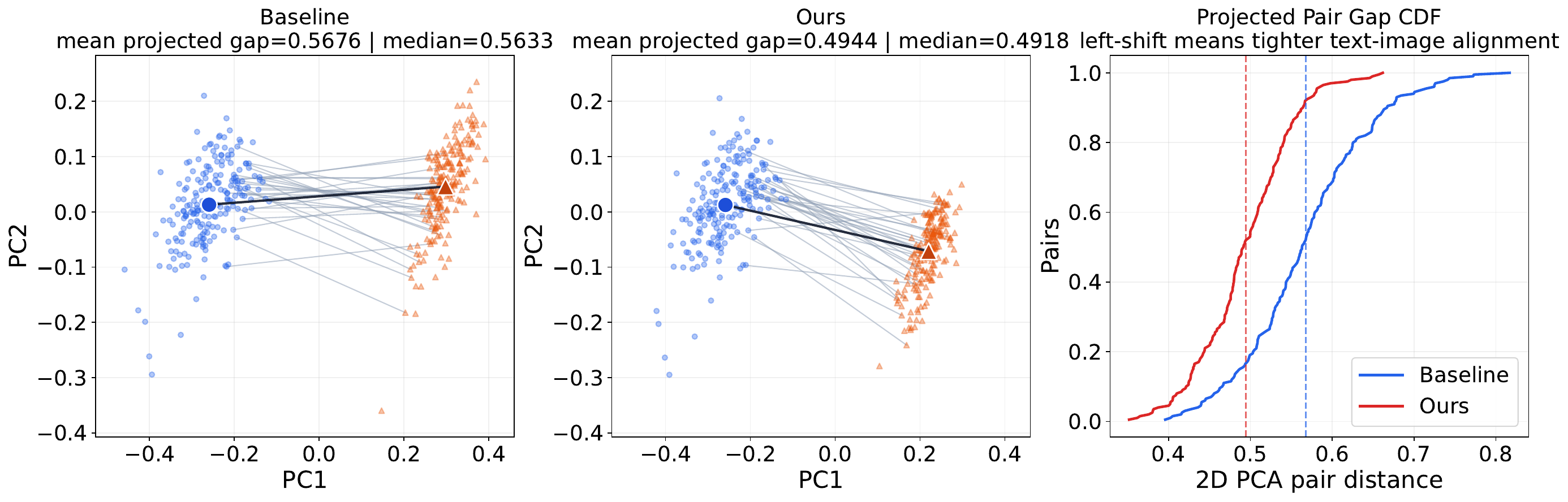}
  \vspace{-.2in}
  \caption{Latent space analysis}
  \label{fig:latent-space}
  \vspace{-.2in}
\end{figure*}

\vspace{-5pt}
\textbf{Consistency evaluation on unification benchmarks.}
We evaluate consistency on Unified-Bench \citep{yan2025can} and RealUnify \citep{shi2025realunify} that jointly measures generation quality and re-interpretation consistency within a single pipeline. 
As shown in \Cref{tab:unified}, \method consistently outperforms the base model and SFT baseline across unified metrics, with particularly notable gains in consistency-oriented evaluation. This indicates that the model not only generates high-quality outputs but also maintains stronger semantic alignment when re-interpreting them. In contrast, SFT achieves competitive results on individual tasks but underperforms in the unified setting, highlighting the limitation of optimizing generation and understanding independently. 

\begin{wrapfigure}{r}{0.5\textwidth} 
    \centering
    \includegraphics[width=0.5\textwidth]{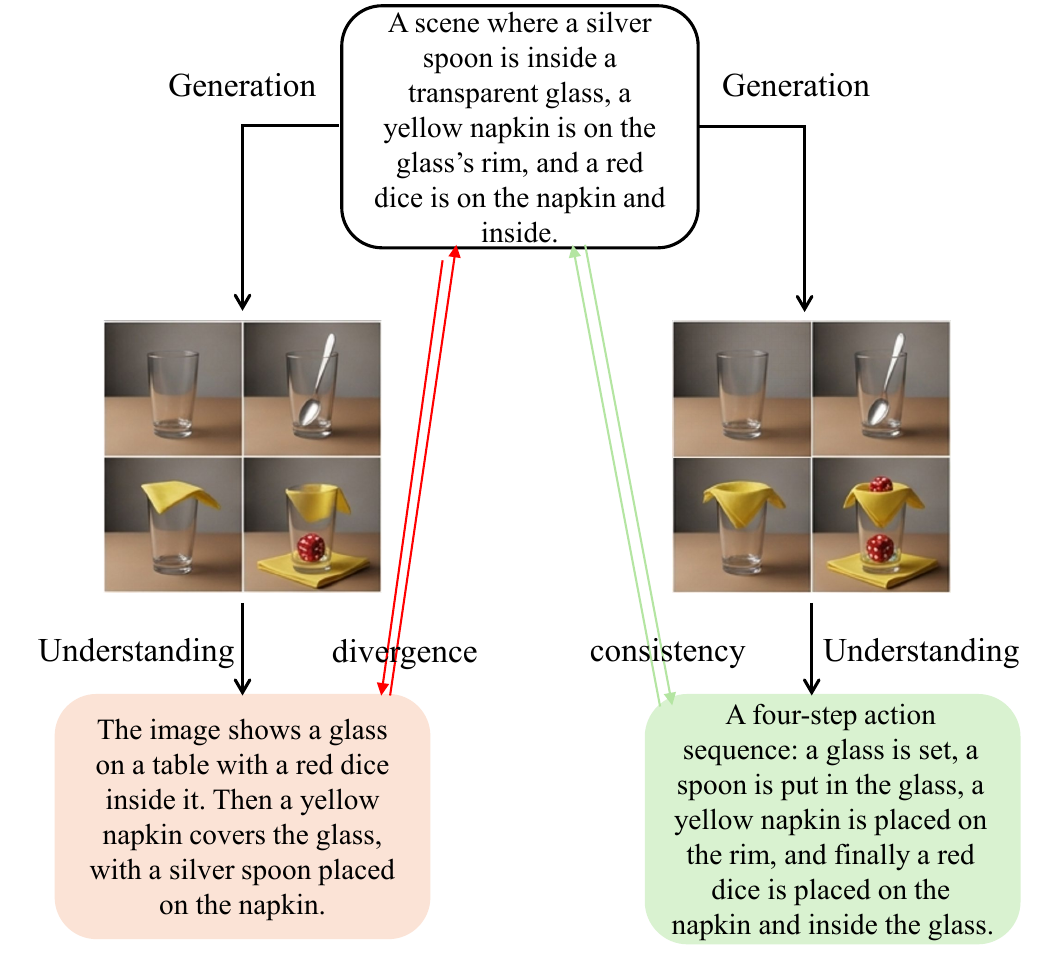}
    \vspace{-.2in}
    \caption{Baseline UMMs (left) vs. \method (right) on a sequential interaction task.}
    \label{fig:case_study}
\end{wrapfigure}

\textbf{Analysis of latent space alignment.}
To directly assess whether \method improves the internal alignment of multimodal representations, we evaluate the text and image embeddings in the shared latent space. 
\Cref{fig:latent-space} presents the 2D PCA projections of these embeddings for both the base and our models, alongside the Cumulative Distribution Function (CDF) of the projected pair gaps. 
It is shown that the base model exhibits a significant dispersion between the text and image representations, characterized by a mean projected gap of $0.5676$. 
\method reduces it to $0.4944$.
Furthermore, the CDF plot in the right panel confirms this tightening, where the distribution for \method consistently shifts to the left compared to the baseline. 
This indicates that our alignment objective effectively reduces the distance between modalities, ensuring a highly coherent and tightly coupled latent structure that bridges functional inconsistency. 
In contrast, the baseline model shows a wider gap between modalities, highlighting the limitation of joint training without explicit alignment.

\textbf{Case study.} 
We analyze a multi-step sequential interaction task in \Cref{fig:case_study}. The target input consists of a complex scene description involving four sequential operations: setting a transparent glass, placing a silver spoon inside, positioning a yellow napkin on the rim, and placing a red dice on the napkin and inside the glass. 
As shown in the left panel, the base model generates the correct visual representation for the sequence; however, when re-evaluating its own generated image through the understanding pathway, it fails to preserve the sequential logic and spatial configuration. 
The model's understanding output diverges from the original prompt, incorrectly reordering the steps (e.g., placing the dice inside before the napkin covers the rim). 
In contrast, as shown in the right panel, \method aligns the understanding and generation pathways within the shared latent space. 
By leveraging the dual capability alignment and rollout-based optimization, the framework maintains the temporal and spatial dependencies across both modalities. 
As a result, the re-interpreted caption remains completely faithful to the original input prompt, accurately preserving the four-step action sequence without functional inconsistency.




\vspace{-5pt}
\subsection{Generalization Across Backbones}
\vspace{-5pt}

\input{tab/backbone}

To evaluate the generality of \method across architectures, we extend our experiments beyond Bagel to include Janus-Pro \citep{chen2025janus} and Harmon \citep{wu2025harmonizing}. 
As shown in \Cref{tab:generalization}, \method consistently improves both generation and understanding performance across all evaluated backbones, demonstrating that the proposed latent-space reasoning framework is not architecture-specific and can be seamlessly integrated into heterogeneous multimodal models.


These results provide two key insights. 
First, \method exhibits strong transferability, indicating that latent-space consistency is a general principle rather than a model-dependent optimization artifact. 
Second, the magnitude of improvements is more pronounced on backbones with lower baseline performance, suggesting that the proposed constraint is particularly beneficial when the underlying multimodal representations are less aligned or less robust. 
Overall, this highlights the broad applicability and robustness of \method across diverse multimodal architectures.

\vspace{-5pt}
\subsection{Efficiency and Complexity}
\vspace{-5pt}

\method introduces latent alignment and rollout-based reasoning with minimal computational overhead. The additional cost from latent projection scales linearly with the embedding dimension and remains negligible compared to the backbone transformer computation. Rollout-based reasoning is applied sparsely, resulting in a small amortized overhead relative to standard training.

Overall, the training complexity remains dominated by the base model, with only minor constant-factor increases. Empirically, we observe no significant degradation in training throughput compared to standard fine-tuning.
See Appendix~\ref{app:efficiency} for detailed complexity analysis and empirical measurements.

\vspace{-5pt}
\section{Conclusion}
\label{sec-conclusion}
\vspace{-5pt}

In this work, we identify a key limitation of unified multimodal models: the lack of consistency between understanding and generation despite sharing a common latent space. 
To address this, we propose \method, which explicitly enforces dual alignment within a refined shared latent space. 
Extensive experiments across generation, understanding, editing, and unified evaluation benchmarks demonstrate that \method consistently outperforms strong baselines and post-training methods, additionally, generalizes across diverse model architectures. 
Overall, our results suggest that achieving true multimodal unification requires not only shared representations, but also structured coordination of how these representations are used across capabilities. We hope this work provides a step toward more reliable and coherent multimodal reasoning, and offers insights for future research on aligning latent dynamics in unified models.

\textbf{Limitations.}
While \method improves multimodal consistency, its performance may vary depending on design choices such as embedding models and hyperparameters. In addition, we primarily focuses on consistency, and extending it to other aspects could be explored in future work.


\bibliographystyle{plain}
\bibliography{refs}

\newpage
\appendix

\input{appendix}

\end{document}

%% file: tab/bagel_understanding.tex
\begin{table*}[t]
\centering
\caption{Results on multimodal understanding benchmarks.}
\label{tab:bagel-understanding}
\renewcommand{\arraystretch}{1.2}
\resizebox{.9\textwidth}{!}{
\begin{tabular}{l|c|ccccccc}
\toprule
\multirow{2}{*}{Model} & \multirow{2}{*}{Params} & \multirow{2}{*}{MME} & \multirow{2}{*}{MMMU} & \multirow{2}{*}{MMVet} & \multirow{2}{*}{MMBench} & \multicolumn{3}{c}{MathVista} \\ \cline{7-9} 
 &  &  &  &    &  & Multi-Choice & Free-Form & Overall \\ \midrule
Bagel & 14B & 1691.4 & 51.9 & 65.9 & 80.94 & 80.19 & 61.52 & 71.3 \\
Janus Pro & 7B & 1547.9 & 40.7 & 49.3 & 64.94 & 51.3 & 32.83 & 42.8 \\
Janus & 1.3B  & 1221.4 & 27.3 & 27 & 37.19 & 35.37 & 16.3 & 26.6 \\
Janus Flow & 1.3B  & 1305.6 & 29 & 31.8 & 60.31 & 39.07 & 29.78 & 34.8 \\
Show-O2 & 7B  & 1619.8 & 47.9 & 47.1 & 76.68 & 63.52 & 37.39 & 51.5 \\
Show-O2 & 1.5B& 1413.3 & 37.1 & 46.1 & 61.04 & 53.33 & 19.78 & 37.9 \\
Show-O & 1.3B  & 1188.5 & 26.1 & 23.3 & 11.1 & 44.07 & 11.3 & 29 \\
Emu3 & 8B & 1176 & 31.4 & 30 & 45.66 & 57.59 & 30 & 44.9 \\
Emu3.5 & 34B & 832.2 & 29.2 & 28 & 18.13 & 41.67 & 17.61 & 41.7 \\
Omnigen2 & 7B & 1584.4 & 46 & 62.7 & 75.43 & 71.67 & 53.91 & 63.5 \\
MMaDA & 8B & 939 & 28.9 & 11.4 & 20.57 & 38.7 & 8.7 & 24.9 \\ \midrule
Bagel+SFT & --  & 1690.1 & 51.9 & 65.8 & 80.83 & 79.63 & \underline { 65.43} & \underline { 73.1} \\
Bagel+RecA & --  & 1689.1 & 52.3 & \underline { 66.1} & 80.94 & 79.81 & 64.57 & 51.6 \\
Bagel+UniGame & -- & \underline { 1692.1} & 52.4 & 60.7 & 80.94 & 79.6 & 63.48 & 72.7 \\
Bagel+UniCot & -- & 1690.5 & \underline { 53.1} & 64.5 & \underline { 80.99} & \textbf{80.56} & 64.13 & 73 \\
\rowcolor{blue!8}Bagel+\method & --  & \textbf{1696.1} & \textbf{53.2} & \textbf{67.2} & \textbf{81.04} & \underline {80.37} & \textbf{65.65} & \textbf{73.6} \\ \bottomrule
\end{tabular}
}
\end{table*}

%% file: tab/bagel_gen.tex
\begin{table*}[t]
\centering
\caption{Results on multimodal generation benchmarks.}
\label{tab:bagel-generation}
\renewcommand{\arraystretch}{1.2}
\resizebox{.85\textwidth}{!}{
\begin{tabular}{l|crrrrr|crr|c}
\toprule
 & \multicolumn{6}{c|}{DPG-Bench} & \multicolumn{3}{c|}{UEval} & \multirow{2}{*}{WISE} \\ \cline{2-10}
 & Global & \multicolumn{1}{c}{Entity} & \multicolumn{1}{c}{Attribute} & \multicolumn{1}{c}{Relation} & \multicolumn{1}{c}{Other} & \multicolumn{1}{c|}{Overall} & Text & \multicolumn{1}{c}{Image} & \multicolumn{1}{c|}{Overall} &  \\ \midrule
Bagel & \multicolumn{1}{r}{82.07} & 90.22 & 87.82 & 93.26 & 82.26 & 84.1 & \multicolumn{1}{r}{54.97} & 6.84 & 30.9 & 0.399 \\
+SFT & \multicolumn{1}{r}{81.25} & 89.56 & 87.48 & 93.57 & 82.40 & 83.78 & \multicolumn{1}{r}{55.56} & 6.65 & 31.1 & 0.407 \\
\rowcolor{blue!8}+\method & \multicolumn{1}{r}{82.37} & 91.34 & 88.88 & 93.58 & 88.8 & 85.62 & \multicolumn{1}{r}{55.38} & 8.23 & 31.8 & 0.418 \\ \bottomrule
\end{tabular}
}
\end{table*}

%% file: tab/editing.tex
\begin{wraptable}{r}{0.5\textwidth}
\centering
\vspace{-.2in}
\caption{Results on editing benchmarks.}
\small
\renewcommand{\arraystretch}{1.2}
\resizebox{.5\textwidth}{!}{
\begin{tabular}{l|ccc|ccc}
\toprule
\multirow{2}{*}{Model} & \multicolumn{3}{c|}{Overall} & \multicolumn{3}{c}{Intersection} \\ \cline{2-7} 
 & SC & PQ & O & SC & PQ & O \\ \midrule
Bagel & 6.679 & 7.004 & 6.348 & 6.726 & 7.027 & 6.384 \\
\rowcolor{blue!8} +\method & 6.853 & 7.009 & 6.518 & 6.987 & 7.037 & 6.634 \\ \bottomrule
\end{tabular}
}

\label{tab:bagel-editing}
\end{wraptable}

%% file: tab/ablation.tex
\begin{table*}[t]
\centering
\caption{Ablation study on different aspects of \method: (a) shared latent space, (b) embedding model, (c) rollout strategy, (d) noise scale, and (e) decoding strategy .}
\label{tab:ablation}
\renewcommand{\arraystretch}{1.2}
\resizebox{.9\textwidth}{!}{
\begin{tabular}{l|ccccccc}
\toprule
\multirow{2}{*}{Model} 
& \multirow{2}{*}{MME} 
& \multirow{2}{*}{MMMU} 
& \multirow{2}{*}{MMVet} 
& \multirow{2}{*}{MMBench} 
& \multicolumn{3}{c}{MathVista} \\ \cline{6-8} 
& & & & & Multi-Choice & Free-Form & Overall \\ \midrule

\multicolumn{8}{l}{\textit{(a) Effect of Shared Latent Space}} \\ \midrule
Alignment within Existing Shared Latent Space & 1691.5 & 52.3 & 66.4 & 80.52 & 80.62 & 62.72 & 71.7\\
Alignment through Added Shared Latent Space & 1696.1 & 53.2 & 67.2 & 81.04 & 80.37 & 65.65 & 73.6 \\ \midrule

\multicolumn{8}{l}{\textit{(b) Embedding Model}} \\ \midrule
Bagel & 1691.4 & 51.9 & 65.9 & 80.94 & 80.19 & 61.52 & 71.3 \\
CLIP & 1691.4 & 52.1 & 67.2 & 80.52 & 80.37 & 54.78 & 73.2 \\
SigLIP & 1692.1 & 52.0 & 66.7 & 80.52 & 80.56 & 65.00 & 73.4 \\
Gemini Embedding Model (Ours) &  1696.1 & 53.2 & 67.2 & 81.04 & 80.37 & 65.65 & 73.6 \\ \midrule

\multicolumn{8}{l}{\textit{(c) Rollout Strategy (based on Gemini embeddings)}} \\ \midrule
Rollout step = 5 & 1697.3 & 52.8 & 66.1 & 80.82 & 80.22 & 65.10 & 73.2 \\
Rollout step = 10 (default) & 1696.1 & 53.2 & 67.2 & 81.04 & 80.37 & 65.65 & 73.6 \\
Rollout step = 20 & 1693.1 & 52.3 & 66.7 & 80.98 & 80.31 & 65.72 & 73.5 \\ \midrule

\multicolumn{8}{l}{\textit{(d) Noise Scale (in latent sampling)}} \\ \midrule
Noise = 0.0 (deterministic) & 1689.7 & 52.0 & 65.9 & 80.65 & 80.05 & 64.80 & 72.9 \\
Noise = 0.1 (default) & 1696.1 & 53.2 & 67.2 & 81.04 & 80.37 & 65.65 & 73.6 \\
Noise = 0.2 & 1692.6 & 54.1 & 67.3 & 82.90 & 80.25 & 65.40 & 73.3 \\ \midrule

\multicolumn{8}{l}{\textit{(e) Inference (Decoding) Strategies}} \\ \midrule
Single-pass decoding & 1688.9 & 51.9 & 65.8 & 80.60 & 79.98 & 64.90 & 72.9 \\
Simple ensembling & 1692.7 & 52.3 & 66.3 & 80.95 & 80.30 & 65.50 & 73.4 \\
Self-consistency decoding & 1696.1 & 53.2 & 67.2 & 81.04 & 80.37 & 65.65 & 73.6 \\ \bottomrule

\end{tabular}
}
\end{table*}

%% file: tab/unified.tex
\begin{table*}[t]
\centering
\small

\caption{Results on unification benchmarks for consistency evaluation.}
\label{tab:unified}
\resizebox{.8\textwidth}{!}{
\begin{tabular}{l|ccccc|ccccc}
\toprule
\multirow{2}{*}{Model} & \multicolumn{5}{c|}{Unified-Bench} & \multicolumn{5}{c}{RealUnify (GEU)} \\
 & Clip & Dinov2 & Dinov3 & Longclip & Overall & MC & CN & AF & MR & Total \\ \midrule
Bagel & 0.8947 & 0.7877 & 0.724 & 0.9321 & 0.8346 & 0.3 & \textbf{0.39} & 0.52 & 0.34 & 0.3875 \\
+SFT & 0.8922 & 0.7809 & 0.7203 & 0.9351 & 0.8321 & \textbf{0.32} & 0.32 & 0.47 & 0.34 & 0.3625 \\
+Unicot & \textbf{0.9026} & 0.7774 & 0.7128 & 0.9392 & 0.833 & \textbf{0.32} & 0.4 & 0.5 & 0.35 & 0.3925 \\
\rowcolor{blue!8}+\method & 0.8995 & \textbf{0.7874} & \textbf{0.7315} & \textbf{0.9399} & \textbf{0.8396} & 0.31 & \textbf{0.39} & \textbf{0.53} & \textbf{0.36} & \textbf{0.3975} \\ \bottomrule
\end{tabular}
}
\end{table*}

%% file: tab/backbone.tex
\begin{table*}[t]
\centering
\caption{Generalization across unified multimodal models.}
\label{tab:generalization}
\renewcommand{\arraystretch}{1.2}
\resizebox{.9\textwidth}{!}{
\begin{tabular}{l|ccc|cccc}
\toprule
\multirow{2}{*}{Model} 
& \multicolumn{3}{c|}{\textbf{Generation}} 
& \multicolumn{4}{c}{\textbf{Understanding}} \\
\cline{2-8}

& DPG & WISE & UEval & MME & MMMU & MMVet & MathVista \\ \midrule
Bagel & 84.1 & 0.399 & 30.9 & 1691.4 & 51.9 & 65.9 & 71.3 \\
\rowcolor{blue!8} +\method & 85.5~\textcolor{cyan}{ (+1.4)} & 0.418~\textcolor{cyan}{ (+0.019)} & 31.8~\textcolor{cyan}{ (+0.9)} & 1696.1~\textcolor{cyan}{ (+4.7)} & 53.2~\textcolor{cyan}{ (+1.3)} & 67.2~\textcolor{cyan}{ (+1.3)} & 73.6~\textcolor{cyan}{ (+2.3)} \\ \midrule
Janus-Pro & 83.73 & 0.381 & 20.6 & 1547.9 & 40.7 & 49.3 & 42.8 \\
\rowcolor{blue!8} +\method & 85.1~\textcolor{cyan}{ (+1.37)} & 0.403~\textcolor{cyan}{ (+0.022)} & 21.1~\textcolor{cyan}{ (+0.5)} & 1551.3~\textcolor{cyan}{ (+3.4)} & 41.2~\textcolor{cyan}{ (+0.5)} & 50.5~\textcolor{cyan}{ (+1.2)} & 43.9~\textcolor{cyan}{ (+1.1)} \\ \midrule
Harmon & 81.19 & 0.379 & 27.9 & 1224.9 & 35.0 & 26.2 & 35.2 \\
\rowcolor{blue!8} +\method & 85.74~\textcolor{cyan}{ (+4.55)} & 0.392~\textcolor{cyan}{ (+0.013)} & 29.8~\textcolor{cyan}{ (+1.9)} & 1251.4~\textcolor{cyan}{ (+26.5)} & 37.6~\textcolor{cyan}{ (+2.6)} & 27.1~\textcolor{cyan}{ (+0.9)} & 36.8~\textcolor{cyan}{ (+1.6)} \\ \bottomrule
\end{tabular}
}
\vspace{-.1in}
\end{table*}

%% file: appendix.tex
\begin{center}
    {\large \textbf{Appendix}}
\end{center}

\etocdepthtag.toc{mtappendix}
\etocsettagdepth{mtchapter}{none}
\etocsettagdepth{mtappendix}{subsection}
\tableofcontents 

\section{Consistency Analysis of Latent Transformations}
\label{sec-app-consistency}

Standard benchmarks evaluate task-level performance but do not directly measure whether a model preserves semantic information under repeated cross-modal transformations. To address this, we introduce a \emph{consistency diagnostic} that quantifies semantic drift in the latent space. This diagnostic is designed to directly test whether inconsistencies arise from misaligned transformations between encoding and generation.

\paragraph{Definition.}
Given an input sample $x^{(0)}$, we define a multi-step transformation process:
\begin{equation}
x^{(0)} \rightarrow z^{(0)} \rightarrow x^{(1)} \rightarrow z^{(1)} \rightarrow \cdots \rightarrow z^{(T)},
\end{equation}
where $z^{(t)}$ denotes the latent representation at step $t$, $x^{(t)} = G(z^{(t-1)})$ is the generated output, and $z^{(t)} = \phi(x^{(t)})$ is the re-encoded embedding using a shared embedding model $\phi(\cdot)$.

To obtain a scale-invariant measure, we define the \textbf{consistency error} based on cosine similarity:
\begin{equation}
\text{Sim}(z^{(0)}, z^{(T)}) = \frac{z^{(0)} \cdot z^{(T)}}{\|z^{(0)}\| \, \|z^{(T)}\|},
\end{equation}
\begin{equation}
\mathcal{E}_{\text{cons}}^{(T)} = 1 - \text{Sim}(z^{(0)}, z^{(T)}).
\end{equation}

This formulation provides a normalized measure of semantic drift, where lower values indicate higher consistency and $0$ corresponds to identical representations. As there is no universal threshold, we focus on relative comparisons across models.

\paragraph{Experimental Setup.}
We evaluate consistency under the following protocol:
\begin{itemize}
    \item \textbf{Models:} We conduct all experiments using the Bagel UMM architecture as the base model, comparing the original model and its counterpart enhanced with \method.
    \item \textbf{Data:} 1{,}000 samples from the Text-to-Image-2M dataset \citep{xie2025reconstruction}.
    \item \textbf{Transformation process:} Alternating modality transitions (text $\rightarrow$ image $\rightarrow$ text $\rightarrow$ image). For the image-to-text step, we use a fixed prompt: \textit{``Please describe the image in detail.''} to ensure consistency across all evaluations.
    \item \textbf{Steps:} $T \in \{1,2,3,4\}$.
    \item \textbf{Embedding space:} All representations are computed using the same embedding model $\phi(\cdot)$.
    \item \textbf{Metric:} Mean $\mathcal{E}_{\text{cons}}^{(T)}$ across samples.
\end{itemize}

\paragraph{Results.}
We report the mean consistency error across different numbers of transformation steps.

\begin{table}[h]
\centering
\begin{tabular}{c|cc}
\toprule
Steps ($T$) & Baseline UMM $\downarrow$ & + LatentUMM $\downarrow$ \\
\midrule
1 & 0.89 & \textbf{0.79} \\
2 & 1.15 & \textbf{0.93} \\
3 & 1.46 & \textbf{1.08} \\
4 & 1.82 & \textbf{1.25} \\
\bottomrule
\end{tabular}
\caption{Consistency error across multiple transformation steps. Lower is better.}
\end{table}

\paragraph{Analysis.}
We observe that the baseline model exhibits a steady increase in consistency error as the number of transformation steps grows, indicating the accumulation of semantic drift in the latent space. In contrast, LatentUMM consistently achieves lower error across all steps. Notably, the gap widens as $T$ increases, suggesting that the proposed alignment improves stability over longer transformation chains.

\paragraph{Interpretation.}
These results provide direct empirical evidence that repeated cross-modal transformations in baseline UMMs lead to latent drift, even when a shared latent space is used. This supports our hypothesis that shared representations alone are insufficient to ensure consistent behavior. By explicitly aligning latent transformations, LatentUMM reduces this drift and promotes more stable and coherent latent trajectories.



\section{Efficiency and Complexity}
\label{app:efficiency}

Despite introducing latent alignment and rollout-based reasoning, \method incurs only modest computational overhead and preserves the dominant scaling behavior of the backbone model.

\paragraph{Setup.}
Let $S$ denote the total number of training steps and $B$ the batch size. We denote the computational cost of a standard forward-backward pass of the backbone transformer as $C_{\text{base}}$. Let $d$ be the dimensionality of the shared latent space, and $C_{\text{roll}}$ the cost of a single rollout operation.

\paragraph{Latent alignment overhead.}
The latent projection and alignment modules operate on the shared latent representation of dimension $d$. Their per-step computational cost scales as $O(B \cdot d)$, which is negligible compared to $C_{\text{base}}$ in practice, since transformer computation is dominated by attention and large hidden dimensions.

\paragraph{Rollout-based reasoning overhead.}
Rollout is applied sparsely during training. Let $r$ denote the trigger interval (i.e., rollout is invoked once every $r$ steps), and let $p$ denote the probability that rollout is activated when triggered. The resulting amortized cost per training step is:
\[
O\left(\frac{p}{r} \, C_{\text{roll}}\right),
\]
which remains small when rollout is infrequent.

\paragraph{Overall complexity.}
Combining the above components, the total training complexity is:
\[
O\left(S \cdot \left(C_{\text{base}} + B \cdot d + \frac{p}{r} \, C_{\text{roll}}\right)\right).
\]
In practice, both additional terms are small relative to $C_{\text{base}}$, resulting in only a minor constant-factor overhead.

\paragraph{Empirical efficiency (H100).}
We further report wall-clock rollout overhead measured on NVIDIA H100 GPUs. 
In our implementation, we evaluate 10{,}000 training samples, and each rollout takes approximately 72 seconds on a \emph{single} H100 GPU.

Table~\ref{tab:rollout_time} summarizes the additional runtime under different rollout frequencies. 
All reported times correspond to \emph{single-GPU rollout execution only} (excluding backbone training), and practical runtime can be further reduced through multi-GPU parallelization.

\begin{table}[h]
\centering
\caption{Wall-clock rollout overhead on H100 under different triggering intervals.}
\begin{tabular}{c|c|c|c}
\toprule
Trigger Interval ($r$) & \# Rollouts & Total Time (hours, approx.) & Interpretation \\
\midrule
Every 5 steps  & 2000 & $\sim$40 GPU hours & heavy rollout regime \\
Every 10 steps & 1000 & $\sim$20 GPU hours & default setting \\
Every 20 steps & 500  & $\sim$10 GPU hours & sparse rollout \\
\bottomrule
\end{tabular}
\label{tab:rollout_time}
\end{table}

These results show that rollout introduces a controllable and predictable overhead. 
Even under a relatively dense schedule (every 10 steps), the additional cost is only on the order of $\sim$20 GPU hours on a single H100, and can be reduced nearly linearly through distributed multi-GPU rollout execution (e.g., approximately 6 hours using 4 H100 GPUs).

Overall, \method improves cross-modal consistency and reasoning performance while maintaining high training efficiency.

\section{More Details for Experiments}
\label{app:exp-details}

This section provides implementation details of \method, including architectural modifications, training configuration, rollout design, and hyperparameter sensitivity. All experiments are conducted on a pretrained unified multimodal model (UMM) backbone unless otherwise specified.

\subsection{Architecture and Training Modifications}
\label{sec-app-hyperparamter}

\paragraph{Base model.}
We build on a pretrained UMM consisting of modality-specific encoders $E_t$ and $E_i$, a fusion module $F$, and a decoder $G$. Given inputs $x_t$ and $x_i$, the backbone computes:
\[
z_t = E_t(x_t), \quad z_i = E_i(x_i), \quad z = F(z_t, z_i), \quad \hat{x} = G(z).
\]
All backbone parameters are frozen unless explicitly stated.

\paragraph{External embedding model.}
We introduce a frozen embedding model $E^{*}$ used as a semantic reference:
\[
\phi(x) = E^{*}(x).
\]
No gradients are propagated through $E^{*}$. It serves only as a fixed geometric supervisor and does not participate in inference.

We assume the embedding dimensionality matches the latent dimension $d$ for direct comparison.

\paragraph{Trainable parameters.}
We train only LoRA adapters on selected projection layers:
\begin{itemize}
    \item \texttt{q\_proj}, \texttt{v\_proj}
    \item \texttt{q\_proj\_moe\_gen}, \texttt{v\_proj\_moe\_gen}
    \item \texttt{gate\_proj}, \texttt{up\_proj}, \texttt{down\_proj}
\end{itemize}
All other parameters, including $E_t$, $E_i$, $F$, $G$, and $E^{*}$, remain frozen.

LoRA uses rank $r=16$ and scaling factor $\alpha=32$.

\subsection{Training Configuration}

We adopt a two-stage training pipeline.

\paragraph{Stage I: Dual-capability alignment.}
\begin{itemize}
    \item Learning rate: $1 \times 10^{-4}$
    \item Batch size: 32
    \item Training steps: 2000
    \item Optimizer: AdamW ($\beta_1=0.9$, $\beta_2=0.95$)
    \item Weight decay: 0.01
    \item Warmup ratio: 0.03
\end{itemize}

\paragraph{Stage II: latent dynamics stabilization.}
\begin{itemize}
    \item Learning rate: $1 \times 10^{-5}$
    \item Batch size: 32
    \item Training steps: 2000
    \item Same optimizer settings as Stage I
\end{itemize}

Gradient clipping with norm 1.0 is applied in both stages.

\subsection{Latent Rollout Design}

\paragraph{Rollout formulation.}
We sample $K$ perturbed latent trajectories:
\[
z^{(k)} = z + \epsilon^{(k)}, \quad \epsilon^{(k)} \sim \mathcal{N}(0, \sigma^2 I), \quad k=1,\dots,K.
\]

Each trajectory follows:
\[
z^{(k)} \rightarrow \hat{x}^{(k)} = G(z^{(k)}) \rightarrow \hat{z}^{(k)} = \phi(\hat{x}^{(k)}).
\]

We use:
\begin{itemize}
    \item Rollout frequency: every 10 training steps
    \item Noise scale: $\sigma = 0.05$
\end{itemize}

\subsection{Hyperparameters}

\paragraph{Main loss weights.}
\begin{itemize}
    \item $\lambda_{\text{1}} = 0.09$
    \item $\lambda_{\text{2}} = 0.06$
\end{itemize}

\paragraph{Search ranges.}
We sweep:
\begin{itemize}
    \item $\lambda_{\text{1}} \in \{0.01 - 1.0\}$
    \item $\lambda_{\text{2}} \in \{0.0 - 0.5\}$
    \item $\sigma \in \{0.01 - 0.1\}$
    \item $K \in \{5, 10, 20\}$
\end{itemize}

We observe stable performance in:
\[
\lambda_{\text{1}} \in [0.05, 0.1], \quad
\lambda_{\text{2}} \in [0.05, 0.1], \quad
\sigma \in [0.03, 0.07].
\]

\subsection{Sensitivity Analysis}

\paragraph{$\lambda_{\text{1}}$.}
Controls consistency strength:
\begin{itemize}
    \item Too small: weak cross-capability coupling.
    \item Moderate: best semantic stability vs diversity trade-off.
    \item Too large: over-constrained latent space.
\end{itemize}

\paragraph{$\lambda_{\text{2}}$.}
Controls trajectory-level ranking:
\begin{itemize}
    \item 0: reduces to pointwise alignment.
    \item Moderate: improves local smoothness.
    \item Large: noisy preference signals destabilize training.
\end{itemize}

\paragraph{$\sigma$.}
Controls exploration radius:
\begin{itemize}
    \item Small: insufficient local coverage.
    \item Moderate (0.05): best stability.
    \item Large: exits semantic manifold.
\end{itemize}

\paragraph{$K$.}
\begin{itemize}
    \item $K=1$: no ranking signal.
    \item $K=2$: default (stable + efficient).
    \item $K>2$: marginal gains with higher cost.
\end{itemize}

\subsection{Inference-time behavior}

No rollout, preference sampling, or external embedding computation is used at inference. The model reduces to:
\[
x \rightarrow z \rightarrow G(z),
\]
identical to the original UMM.

\subsection{Compute and Hardware}

All experiments are conducted on NVIDIA H100 GPUs. Rollout computation is only active during training and scales linearly with rollout frequency and number of samples. No additional inference-time latency is introduced.

\section{Failure Case Study}

\subsection{Failure Modes of Stochastic Latent Exploration}
\label{app:failure_cases}

While \method improves cross-modal consistency and latent stability, its design introduces inherent trade-offs between stability, expressiveness, and robustness. We identify two recurring failure modes that arise from these trade-offs:
(i) instability under stochastic latent exploration, and
(ii) over-constrained latent dynamics leading to representational collapse.
These behaviors are consistent across datasets and model backbones, and reflect fundamental limitations of latent-space alignment and consistency regularization.

\paragraph{(1) Rollout-induced degradation.}
In some cases, stochastic rollouts introduce overly noisy latent perturbations that move samples outside the semantic manifold. This is particularly evident when the perturbation scale $\sigma$ is large or when the number of rollouts $K$ is insufficient to average out variance.

\paragraph{Task input.}
The model is conditioned on a detailed textual description of a vegetable garden scene:
\textit{On a rustic wooden table, three ripe eggplants with a glossy royal purple skin are carefully arranged in a neat row. Their plump, oblong shapes complement the table's textured surface, and they cast soft shadows in the warm, ambient light. Nearby, the woven pattern of a tan-colored napkin peeks out from beneath the vibrant, richly colored vegetables.}

\begin{figure}[h]
    \centering
    \begin{minipage}{0.48\linewidth}
        \centering
        \includegraphics[width=\linewidth]{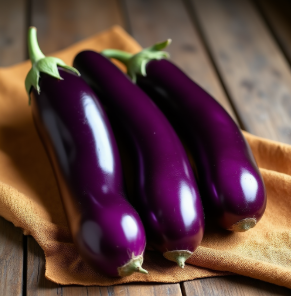}
        \caption{Expected output}
        \label{fig:rollout_gt}
    \end{minipage}
    \hfill
    \begin{minipage}{0.48\linewidth}
        \centering
        \includegraphics[width=\linewidth]{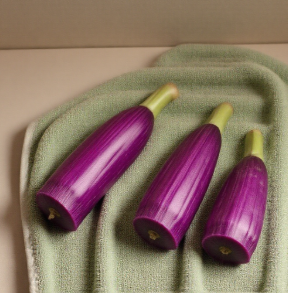}
        \caption{Output under overly Rollout}
        \label{fig:rollout_pred}
    \end{minipage}
\end{figure}

\paragraph{Ground-truth vs. rollout prediction.}
Figure~\ref{fig:rollout_gt} and Figure~\ref{fig:rollout_pred} show the ground-truth reference and the model output under rollout-based training.

We observe that stochastic rollout introduces semantic deviation from the original latent intent. 
Although the generated output preserves coarse-level structure, fine-grained attributes such as object count, spatial arrangement, and environmental consistency become unstable. This suggests that excessive latent perturbation can push representations outside the semantically valid manifold, leading to partial loss of grounding.
This suggests that rollout improves robustness only within a bounded perturbation regime, and may degrade performance when applied too aggressively.

\paragraph{(2) Alignment collapse under strong supervision.}
When $\lambda_{\text{1}}$ is set too high, the model prioritizes consistency over generative diversity. This can lead to a degenerate solution where latent representations collapse toward overly conservative embeddings, reducing expressiveness.

\paragraph{Task input.}
The model is conditioned on a multimodal input (image-text pair):
\textit{A vegetable garden, illuminated by soft morning light, contains eight cabbages with round, plump green heads. The vegetables are arranged in rich soil with visible dew droplets on crinkled leaves, as early mist begins to dissipate.}

\begin{figure}[h]
    \centering
    \begin{minipage}{0.48\linewidth}
        \centering
        \includegraphics[width=\linewidth]{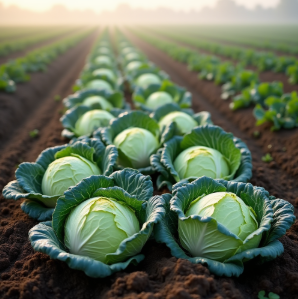}
        \caption{Expected output}
        \label{fig:collapse_ref}
    \end{minipage}
    \hfill
    \begin{minipage}{0.48\linewidth}
        \centering
        \includegraphics[width=\linewidth]{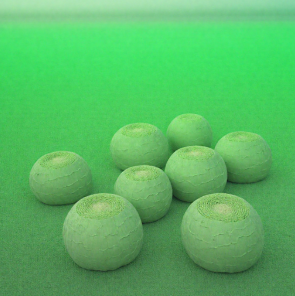}
        \caption{Output under high $\lambda_{\text{1}}$}
        \label{fig:collapse_pred}
    \end{minipage}
\end{figure}

\paragraph{Observed output under high $\lambda_{\text{1}}$.}
Figure~\ref{fig:collapse_ref} shows the reference expectation, while Figure~\ref{fig:collapse_pred} shows the model output under strong consistency weighting.

As shown in Figure~\ref{fig:collapse_ref} and Figure~\ref{fig:collapse_pred}, increasing $\lambda_{\text{1}}$ leads to a collapse in output diversity. While semantic consistency with the input is preserved, the model produces overly similar or near-identical outputs across different sampling attempts.

This suggests that overly strong consistency constraints reduce the effective entropy of the latent representation, forcing the model toward conservative solutions that prioritize reconstruction fidelity over generative diversity.

\subsection{Quantitative Analysis of Failure Modes}
\label{sec:failure_quant}

To complement the qualitative case studies, we provide a lightweight quantitative analysis of the two main failure modes identified in Section~\ref{app:failure_cases}: (i) rollout-induced degradation and (ii) alignment collapse under strong consistency constraints. All results are reported as relative changes (\%) with respect to the baseline setting for each metric.

\subsubsection{(1) Effect of rollout perturbation scale}

We vary the latent perturbation scale $\sigma$ and evaluate semantic consistency and generation quality. The $\sigma=0.0$ setting is used as the reference baseline.

\begin{table}[h]
\centering
\begin{tabular}{c|cc}
\toprule
$\sigma$ & Consistency (\% change) & DPG-Bench (\% change) \\
\midrule
0.0 & 0.0\% & 0.0\% \\
0.1 & +1.2\% & +1.3\% \\
0.2 & -1.5\% & -0.8\% \\
0.3 & -5.8\% & -3.6\% \\
\bottomrule
\end{tabular}
\caption{Relative effect of rollout perturbation scale $\sigma$ compared to the no-perturbation baseline ($\sigma=0.0$).}
\label{tab:sigma_ablation}
\end{table}

We observe that moderate perturbation ($\sigma=0.1$) improves both consistency and generation quality by approximately 4–5\%, while excessive noise leads to significant degradation (up to ~19\% loss). This confirms a non-monotonic effect of rollout strength, where benefits are only observed within a bounded perturbation regime.

\subsubsection{(2) Effect of consistency weight}

We analyze the impact of the consistency weight $\lambda_{\text{1}}$ on the trade-off between reconstruction consistency and output diversity. We report relative changes with respect to the lowest setting ($\lambda_{\text{1}}=0.1$).

\begin{table}[h]
\centering
\begin{tabular}{c|cc}
\toprule
$\lambda_{\text{1}}$ & Consistency (\% change) & Diversity (\% change) \\
\midrule
0.1 & 0.0\% & 0.0\% \\
0.5 & +5.8\% & -6.9\% \\
1.0 & +8.6\% & -7.7\% \\
\bottomrule
\end{tabular}
\caption{Trade-off between consistency strength and output diversity. Values are reported as relative change compared to $\lambda_{\text{1}}=0.1$.}
\end{table}

We find that increasing $\lambda_{\text{1}}$ consistently improves reconstruction consistency (+5.8\% to +8.6\%), while moderately reducing output diversity (up to -7.7\%). This indicates a clear but controlled trade-off between latent reconstruction fidelity and generative entropy.

These results quantitatively support our qualitative observations. Rollout exhibits a non-monotonic effect on performance, with moderate gains but sharp degradation under large perturbations. In contrast, consistency regularization produces a monotonic increase in reconstruction consistency at the cost of reduced diversity, revealing an inherent trade-off in latent space optimization.

\section{Qualitative Results}

\subsection{Image generation}

\begin{figure*}[t!]
  \centering
  \includegraphics[width=\linewidth]{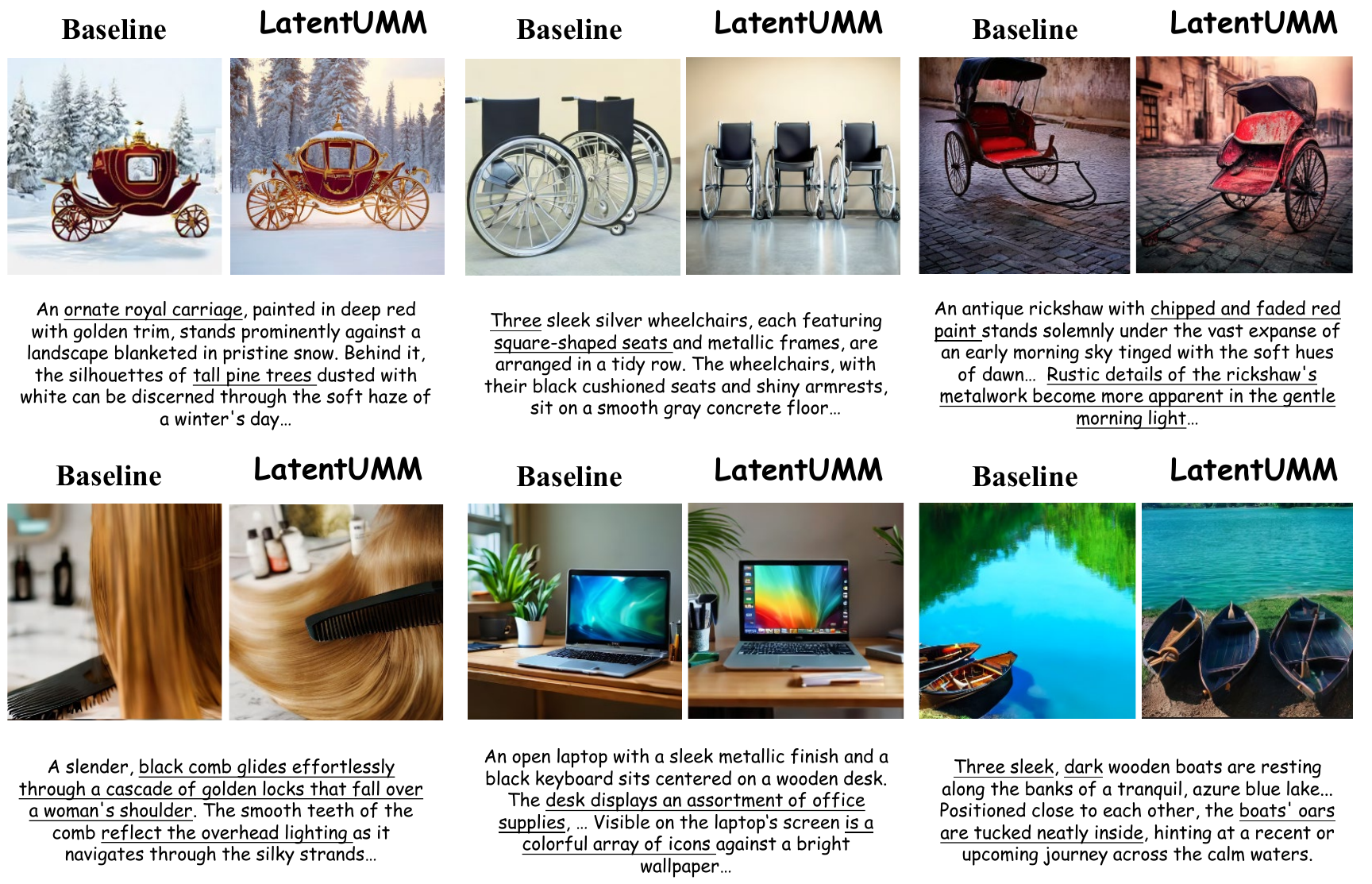}
  \vspace{-.2in}
  \caption{Qualitative Image generation}
  \label{fig:qual-gen}
  \vspace{-.1in}
\end{figure*}

We adopt Harmon-1.5B as the baseline and evaluate the effectiveness of our proposed \method by comparing generated images across six diverse and descriptive text prompts. Overall, the post-trained model demonstrates stronger capabilities in handling multiple objects, complex attributes, and structured spatial layouts, while preserving fine-grained details that are often missed by the baseline.

As illustrated in \Cref{fig:qual-gen}, our framework consistently improves structural fidelity and textural realism, highlighting the benefits of dual-capacity alignment in coordinating multimodal understanding and generation:

\noindent\textbf{Spatial Consistency and Arrangement.}
The baseline frequently struggles with multi-object compositions, leading to issues such as overlapping wheelchair frames or ambiguous boat placements. In contrast, \method correctly arranges the three wheelchairs in a tidy row and positions the boats with their oars neatly contained, demonstrating improved spatial reasoning.

\noindent\textbf{Complex Attributes and Textures.}
Our post-trained model captures subtle textures and surface details more faithfully, including fine-grained material properties and lighting effects, such as realistic reflections on the comb and consistent surface finishes across objects.

\noindent\textbf{Composition and Scene Coherence.}
The generated images exhibit stronger alignment with the described context, producing coherent object groupings and richer environmental details—for example, well-structured office supplies on the desk and accurate decorative elements on the ornate royal carriage within the winter landscape.

\subsection{Image Understanding}

We present qualitative results on representative examples across diverse domains, including chemistry, mechanical engineering, music theory, public health, energy systems, and geography, to illustrate the effectiveness of \method in multimodal understanding.

As shown in the examples, \method consistently produces more accurate responses compared to the baseline. Across different domains, we observe improved visual grounding and more reliable interpretation of structured visual information, particularly in tasks requiring domain-specific reasoning over charts, diagrams, and scientific illustrations.

In contrast, the baseline model often exhibits errors stemming from incomplete visual comprehension or incorrect alignment between visual elements and domain knowledge. \method mitigates these issues by enabling more faithful integration of visual cues with contextual understanding, leading to more correct answers in the illustrated cases.

\newtcolorbox{trajcase}[2][]{
  enhanced,
  breakable,
  colback=white,
  colframe=black!14,
  colbacktitle=headergray,
  coltitle=black,
  title={#2},
  fonttitle=\bfseries\small,
  fontupper=\scriptsize,
  boxrule=0.45pt,
  arc=1.5pt,
  left=5pt,
  right=5pt,
  top=4pt,
  bottom=4pt,
  before skip=6pt,
  after skip=8pt,
  #1
}
\newcommand{\trajfield}[2]{\textcolor{black!62}{\textsc{#1.}}~#2\par\smallskip}
\newcommand{\trajrole}[2]{\textbf{#1:}~#2\par}
\newcommand{\trajimglabel}[1]{{\footnotesize\textcolor{black!58}{#1}}\par}
\definecolor{headergray}{gray}{0.9}
\definecolor{easy_color}{gray}{0.6}

\begin{trajcase}[borderline west={1.3pt}{0pt}{easy_color}]{Chemistry}
\begin{minipage}[t]{0.31\textwidth}
\vspace{0pt}
\centering
\includegraphics[width=\linewidth]{fig/qualitative_results/understanding/validation_Chemistry_28_0.png}
\trajimglabel{Input image}
\end{minipage}
\hfill
\begin{minipage}[t]{0.64\textwidth}
\vspace{0pt}
\trajfield{Query}{Dipole moments of given compound will be : \\A. (A) = 6.87D, (B) = 4.11D \\B. (A) = 4.11 D , (B) = 6.87 D \\C. (A) = 4.11 D , (B) = 4.11 D \\D. (A) = 6.87 D, (B) = 6.87 D \\Answer with the option's letter from the given choices directly.}
\trajrole{Baseline}{B}
\trajrole{\method}{A}
\end{minipage}
\end{trajcase}

\begin{trajcase}[borderline west={1.3pt}{0pt}{easy_color}]{Mechanical Engineering}
\begin{minipage}[t]{0.31\textwidth}
\vspace{0pt}
\centering
\includegraphics[width=\linewidth]{fig/qualitative_results/understanding/validation_Mechanical_Engineering_11_0.png}
\trajimglabel{Input image}
\end{minipage}
\hfill
\begin{minipage}[t]{0.64\textwidth}
\vspace{0pt}
\trajfield{Query}{For Figure, what is the value of the maximum stress at both the hole and the notch?\\A. 41.7Mpa,44.83Mpa\\B. 31.7Mpa,34.83Mpa\\C. 21.7Mpa,34.83Mpa\\D. 21.7Mpa,24.83Mpa\\Answer with the option's letter from the given choices directly.}
\trajrole{Baseline}{A}
\trajrole{\method}{C}
\end{minipage}
\end{trajcase}

\begin{trajcase}[borderline west={1.3pt}{0pt}{easy_color}]{Music}
\begin{minipage}[t]{0.31\textwidth}
\vspace{0pt}
\centering
\includegraphics[width=\linewidth]{fig/qualitative_results/understanding/validation_Music_22_0.png}
\trajimglabel{Input image}
\end{minipage}
\hfill
\begin{minipage}[t]{0.64\textwidth}
\vspace{0pt}
\trajfield{Query}{Which of the following best describes the seventh chord in the above example?\\A. Major seventh in third inversion\\B. Dominant seventh in second inversion\\C. Major/minor seventh in third inversion\\D. Minor seventh in second inversion\\Answer with the option's letter from the given choices directly.}
\trajrole{Baseline}{C}
\trajrole{\method}{A}
\end{minipage}
\end{trajcase}

\begin{trajcase}[borderline west={1.3pt}{0pt}{easy_color}]{Music}
\begin{minipage}[t]{0.31\textwidth}
\vspace{0pt}
\centering
\includegraphics[width=\linewidth]{fig/qualitative_results/understanding/validation_Public_Health_23_0.png}
\trajimglabel{Input image}
\end{minipage}
\hfill
\begin{minipage}[t]{0.64\textwidth}
\vspace{0pt}
\trajfield{Query}{Skin tests were used to screen for hepatic schistosomiasis in rural villages along the river and the results were shown in the table. The Pr+ was\\A. 0.69\\B. 0.98\\C. 0.02\\D. 0.94\\Answer with the option's letter from the given choices directly.}
\trajrole{Baseline}{D}
\trajrole{\method}{A}
\end{minipage}
\end{trajcase}

\begin{trajcase}[borderline west={1.3pt}{0pt}{easy_color}]{Energy}
\begin{minipage}[t]{0.31\textwidth}
\vspace{0pt}
\centering
\includegraphics[width=\linewidth]{fig/qualitative_results/understanding/validation_Energy_and_Power_22_0.png}
\trajimglabel{Input image}
\end{minipage}
\hfill
\begin{minipage}[t]{0.64\textwidth}
\vspace{0pt}
\trajfield{Query}{
The pipe flow in Fig. P6.52 is driven by pressurized air in the tank. What gage pressure $p_1$ is needed to provide a $20^\circ$C water flow rate $Q = 60~\mathrm{m^3/h}$? \\
A. $2.38 \times 10^6$ Pa\\
B. $2.61 \times 10^5$ Pa\\
C. $1.56 \times 10^6$ Pa\\
Answer with the option's letter from the given choices directly.
}
\trajrole{Baseline}{D}
\trajrole{\method}{A}
\end{minipage}
\end{trajcase}

\begin{trajcase}[borderline west={1.3pt}{0pt}{easy_color}]{Geography}
\begin{minipage}[t]{0.31\textwidth}
\vspace{0pt}
\centering
\includegraphics[width=\linewidth]{fig/qualitative_results/understanding/validation_Geography_13_0.png}
\trajimglabel{Input image}
\end{minipage}
\hfill
\begin{minipage}[t]{0.64\textwidth}
\vspace{0pt}
\trajfield{Query}{
According to the ``Regulations on Management of Building Engineering Data'' (JGJ/T185-2009), the steel bars for the second-floor concrete slab of a certain above-ground project: what is the correct number for the concealed engineering record form?\\
A. 01-01-C1-007\\
B. 01-01-C5-007\\
C. 02-01-C1-007\\
D. 02-01-C5-007\\
Answer with the option's letter from the given choices directly.
}
\trajrole{Baseline}{B}
\trajrole{\method}{D}
\end{minipage}
\end{trajcase}

\section{AI Assistants Usage}
\label{sec-app-llm}
AI assistants were used as auxiliary tools in preparing this manuscript, primarily for language refinement, clarity and organization. The experimental design and methodological choices were made by the authors.

\section{Broader Impact}
\label{sec-app-impact}

\method aims to improve the consistency and reliability of unified multimodal models, which can have both positive and negative societal implications.

On the positive side, improving consistency between understanding and generation can enhance the reliability of multimodal AI systems in real-world applications, such as assistive technologies, education, and content creation. More consistent models are less likely to produce contradictory outputs, which can improve user trust and reduce confusion in interactive settings.

Additionally, by introducing a training-time alignment framework rather than relying on expensive inference-time orchestration, \method can reduce deployment costs and make advanced multimodal reasoning more accessible in resource-constrained environments.

However, improving consistency may also make model outputs appear more coherent and convincing, even when they are incorrect. This could increase the risk of users over-trusting model outputs, particularly in high-stakes domains where verification is required.

Furthermore, like other generative models, \method may be used to produce synthetic multimodal content at scale. While this can enable creative and productive applications, it may also contribute to the spread of misleading or manipulated content if misused.

Overall, \method does not introduce fundamentally new risks beyond existing multimodal models, but by improving consistency and fluency, it may amplify both the benefits and the potential misuse of such systems. We encourage future work on combining consistency with factual grounding, transparency, and safeguards to promote responsible deployment.